# Can AutoML outperform humans? An evaluation on popular OpenML datasets using AutoML Benchmark

Marc Hanussek
University of Stuttgart, Institute of Human Factors and Technology Management (IAT), Stuttgart, Germany
Email: marc.hanussek@iat.uni-stuttgart.de

Matthias Blohm
University of Stuttgart, Institute of Human Factors and Technology Management (IAT), Stuttgart, Germany
Email: matthias.blohm@iat.uni-stuttgart.de

Maximilien Kintz
Fraunhofer IAO, Fraunhofer Institute for Industrial Engineering IAO, Stuttgart, Germany
Email: maximilien.kintz@iao.fraunhofer.de

*Abstract*—In the last few years, Automated Machine Learning (AutoML) has gained much attention. With that said, the question arises whether AutoML can outperform results achieved by human data scientists. This paper compares four AutoML frameworks on 12 different popular datasets from OpenML; six of them supervised classification tasks and the other six supervised regression ones. Additionally, we consider a real-life dataset from one of our recent projects. The results show that the automated frameworks perform better or equal than the machine learning community in 7 out of 12 OpenML tasks.

## I. INTRODUCTION

Automated Machine Learning approaches that try to solve regression or classification problems on large datasets are gaining growing popularity and importance. Since they are subject to constant changes and are getting already better even with the automated task of proper feature engineering, the discussion whether AutoML tools will replace human scientists in the future is more and more pressing [12].

There is a lot of recent work that provides benchmarking scores for current state of the art AutoML results. Nonetheless, comparison with the performance that human data scientists could have achieved on the same tasks and datasets by manual feature engineering and model tuning is often missing. Our work targets an approach for closing this gap.

A good starting point to compare performance of human and AutoML tools on specific tasks is given by the OpenML platform, which provides many tasks and datasets together with a leaderboard that also lists all details about the used models and algorithms that achieved the scores [11].

This work first gives an overview of related work in section 2 and AutoML benchmarks, followed by the experimental setup and methodology in section 3. In section 4, we describe our findings on the different tasks and AutoML tools. Section 5 closes this work with a conclusion and an outlook.

## II. RELATED WORK

There have been several recent studies describing the current state of the art of AutoML and available frameworks [6, 9, 13], which already reveal impressive performance of AutoML on many datasets and tasks. Additionally, tasks like the ChaLearn Challenge [8] came up that are targeting AutoML approaches without any human intervention. However, most work is limited to comparison of different AutoML tools on different tasks and not regarding how well a human could solve the task without using AutoML frameworks. Nonetheless, additional important domain knowledge that cannot be mapped to AutoML tools yet in an automated fashion may provide advantages for a human data scientist [4].

More details about AutoML and corresponding tools can be found in the work of [7].

## III. METHODOLOGY OF THE BENCHMARK

As described in Figure 1, our benchmark consists of an evaluation of four AutoML tools, which we tested on 12 data sets coming from OpenML and one resulting from one of our projects. For performing the experiments, we used the open source tool AutoML benchmark [3], which comes with full integration of OpenML datasets for many

AutoML frameworks as well as automated benchmarking functions.

*A. Datasets*

Supervised classification and supervised regression are the most popular machine learning tasks and the most considered tasks on OpenML, hence we chose six of each. In the classification case, we picked the six tasks with the most runs on OpenML. In the regression case, we proceeded similarly with the exception that we exchanged two of the six tasks with most runs as all the runs were submitted by only one user. We explicitly state the considered tasks in the result table.

By considering popular OpenML tasks only, we construct the best human performance as the best result stated on the leaderboard and not indicating usage of an AutoML method.

In addition to the 12 tasks from OpenML, we examined a real-life dataset. This dataset, which is available as a csv file, comprises more than 300000 data points having 35 features (numerical, categorical and Boolean). The task considered is a supervised regression one. Data scientists previously preprocessed the data because of poor data quality and associated non-suitability for AutoML Benchmark (or machine learning algorithms in general).

*B. AutoML Benchmark*

We mainly ran the benchmarks with default settings as defined in config.yaml in the AutoML Benchmark project, i.e. usage of all cores, 2GiB of memory left to the OS, amount of memory computed from os available memory and many more.

The first parameter we set was the runtime per fold, which we set to one hour. Once-only, for the most popular task on OpenML ("supervised classification credit-g"), we allowed the best of the four AutoML frameworks (meaning the AutoML framework performing best on supervised classification credit-g with one hour runtime per fold) a runtime per fold of five hours in order to see if it is now able to beat the best result achieved by humans.

The second parameter we set were the metrics in the supervised regression tasks, which we describe in section 3.4.

*C. AutoML frameworks*

We considered the four AutoML frameworks TPOT [10], H2O [5], auto-sklearn [2] and AutoGluon [1]. On the one hand, this constitutes a mix of very recent

AutoML frameworks and frameworks that have been around a bit longer. On the other hand, the selection encompasses Deep Learning-only AutoML frameworks as well as scikit-learn-based AutoML frameworks as well as AutoML frameworks that make use of both approaches.

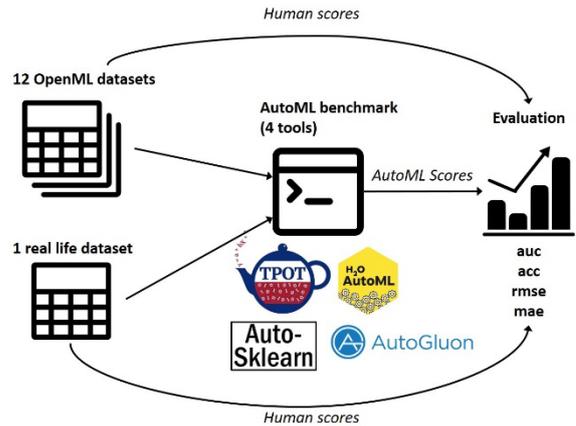

Figure 1. Process workflow of the AutoML benchmarking experiments

TABLE I. OVERVIEW OVER TASK NUMBERS AND CORRESPONDING METRICS

| Task number | Task type | Primary metric | Other metric |
|---|---|---|---|
| 1-6 | Supervised classification | auc | acc |
| 7-12 | Supervised regression | rmse | mae |

*D. Cross-Validation and Metrics*

In order to get reliable results, we required the tasks to have 10-fold cross-validation, which, however, was the case either way. In the supervised classification tasks, we used evaluation methods preconfigured by AutoML Benchmark, which were ROC AUC (auc) and accuracy (acc). For the supervised regression tasks, we chose root-mean-square error (rmse) and mean absolute error (mae), since the metrics preconfigured by AutoML Benchmark included R2, which we were not able to retrieve from the OpenML website. The frameworks optimized for auc and rmse, the so-called primary metrics. The other ones, acc and mae, are stated for informational purposes. An overview over the tasks and the used metrics is given in TABLE I.

*E. Hardware*

The machine we ran the benchmark on was a dedicated server we host locally at Fraunhofer IAO. The server is equipped with two Intel Xeon Silver 4114 CPUs @2.20Ghz (yielding 20 cores in total), four 64GB DIMM DDR4 Synchronous 2666MHz memory modules and two NVIDIA GeForce GTX 1080 Ti (yielding more than 22GB VRAM in total).

## IV. RESULTS AND ANALYSIS

Appendix A shows the main results. Note that in the last two columns we considered only the primary metric. We subsequently state the main results including comments:

1. In 7 out of 12 cases, AutoML performed better or equal than humans in the primary metric.

2. In the same 7 out of 12 cases, AutoML performed better or equal than humans in both metrics.

*Hence, there does not seem to be a significant difference between the primary and the other metric regarding performance.*

3. All these seven cases are either "easy" classification tasks (meaning tasks that humans as well as AutoML solved perfectly) or regression tasks.

*In OpenML (or its community, respectively) regression tasks are not as extensively examined as classification tasks (the six classification tasks with most runs feature in total more than 1.8 million runs while the regression counterparts only have 377 runs). This means that we generally can expect better results for classification tasks in OpenML. In this sense, difference in AutoML performance between non-easy classification tasks on the one hand and regression tasks on the other hand can be easily explained. Nonetheless, AutoML managed to solve the easy classification tasks as perfectly as humans did.*

4. Most results achieved by AutoML are only slightly better or worse than the ones by humans.

*Ignoring the two regression tasks with tremendous AutoML outperformance, the mean of mapes (mean absolute percentage errors) is not even 1% (0.83%). As already mentioned in 3., this might be due to the fact that supervised regression tasks are not as thoroughly considered on OpenML as supervised classification ones.*

5. H2O, the best AutoML framework for supervised classification credit-g, achieves an auc score of 0.7892 using 5h time limit per fold instead of 0.799 using 1h time limit per fold.

*This is unexpected as the train/test splits included in the task are not chosen randomly but are stated explicitly (which can be seen via OpenML API). We assume that the framework internally creates random seeds during model generation that may lead to varying performances.*

6. If AutoML is better or equal in the primary metric, this holds true for three out of four AutoML frameworks on average.

*This shows that, in general, outperformance by AutoML frameworks is a matter of the underlying dataset, strength of the human results or task type rather than the choice of a specific AutoML framework.*

7. The performance of AutoML on the real-life damage events dataset was significantly better than the one achieved by us. More precisely, using AutoML yields a mae 25% better than achieved by conventional methods.

*The performance comparison on the 12 OpenML datasets and the one provided by us is inconsistent in as much as our dataset was preprocessed by humans (which was done in our previous project work) whereas the OpenML ones were not. Therefore, the AutoML frameworks had an advantage on our dataset, which explains, at least partly, the difference in outperformance on this dataset and the 12 OpenML datasets.*

## V. CONCLUSION, OUTLOOK AND FUTURE WORK

Our work shows that using AutoML frameworks constitute a serious strategy when working on machine learning experiments or projects. Partially, they outperform human data scientists and, if not, they often deliver nearly similar results. Future improvements in existing frameworks or the introduction of new frameworks will only support or speed up this trend. In our opinion, usage of AutoML as a general approach when working on machine learning experiments is beneficial for both machine learning beginners and machine learning veterans. Apart from installing the frameworks and interpreting the results, beginners do not need to dive deep into machine learning terms in order to get decent results. Experts can rapidly establish results that define a good baseline, if not more, without spending a lot of time for feature engineering or model tuning.

In our opinion, a future challenge is combining domain knowledge and using AutoML as a general approach in machine learning projects. Domain experts are often equipped with domain knowledge that may prove valuable to machine learning algorithms. Today, it seems that users are not able to incorporate this knowledge into AutoML frameworks appropriately, as these frameworks work highly autonomously. Similar to facilitating the use of machine learning in general, AutoML frameworks could facilitate incorporating domain knowledge by non-specialists.

In future work we want to examine suitability and performance of AutoML approaches and frameworks for text classification on real-life datasets.

### CONFLICT OF INTEREST

The authors declare no conflict of interest.

### AUTHOR CONTRIBUTIONS

Marc Hanussek and Matthias Blohm conducted the research. All three authors wrote the paper together and approved the final version.

APPENDIX A  MAIN EXPERIMENTAL RESULTS

| Number | OpenML task name | OpenML task id | Type | Best human score | | | Best score by AutoML (auc resp. rmse) | AutoML better or equal in at least one metric? | Best AutoML framework outperforms humans by ... % | Number of AutoML frameworks better or equal |
|---|---|---|---|---|---|---|---|---|---|---|
| 1 | supervised classification credit-g | 31 | classification | metric | acc 0.786 | auc 0.807 | 0.799 | NO | -0.97 | 0 |
| 2 | supervised classification blood-transfusion | 10101 | classification | metric | acc 0.8021 | auc 0.756 | 0.754982 | NO | -0.16 | 0 |
| 3 | supervised classification wilt | 9914 | classification | metric | acc 0.9897 | auc 0.996 | 0.995346 | NO | -0.08 | 0 |
| 4 | Supervised classification tic-tac-toe | 145804 | classification | metric | acc 1 | auc 1 | 1 | YES | 0 | 4 |
| 5 | Supervised Classification on monks-problems-2 | 146065 | classification | metric | acc 1 | auc 1 | 1 | YES | 0 | 3 |
| 6 | Supervised Classification on monks-problems-1 | 146064 | classification | metric | acc 1 | auc 1 | 1 | YES | 0 | 4 |
| 7 | Supervised Regression on cholesterol | 2295 | regression | metric | mae 38.6187 | rmse 50.76 | 49.72166 | YES | 2.04 | 4 |
| 8 | Supervised Regression on liver-disorders | 52948 | regression | metric | mae 2.3088 | rmse 2.941 | 2.995219 | NO | -1.83 | 4 |
| 9 | Supervised Regression on analcatdata_negotiation | 4823 | regression | metric | mae 0.5766 | rmse 0.789 | 0.79956 | NO | -1.33 | 0 |
| 10 | Supervised Regression on cleveland | 2285 | regression | metric | mae 0.6407 | rmse 0.873 | 0.856123 | YES | 1.93 | 1 |
| 11 | Supervised Regression on fri_c3_100_25 | 4958 | regression | metric | mae 0.7575 | rmse 0.938 | 0.255819 | YES | 72.73 | 4 |
| 12 | Supervised Regression on kin8nm | 2280 | regression | metric | mae 0.1634 | rmse 0.203 | 0.005561 | YES | 97.26 | 4 |

Marc Hanussek holds a Master's degree in Mathematics from Ulm University (Germany). Since his graduation, he works for Fraunhofer IAO in Stuttgart, a subsidiary institute of Fraunhofer Society, the largest applied research organization in Europe. His work revolves around machine learning with special interest in AutoML and Explainable Artificial Intelligence.

Matthias Blohm holds a Master's degree in Software Engineering from University of Stuttgart (Germany). Since his graduation, he works for Fraunhofer IAO in Stuttgart. His work revolves around deep learning with special interest in Deep Learning and NLP.

Maximilien Kintz holds a PhD in Engineering from University of Stuttgart (Germany). Since his graduation, he works for Fraunhofer IAO in Stuttgart. His work revolves around artificial intelligence and data science with special interest in analysis of unstructured text data and information visualization.